\documentclass{article}

\usepackage{amsmath}
\usepackage{graphicx}
\usepackage{subcaption}

 \usepackage[main, final]{neurips_2026}

\usepackage[utf8]{inputenc} 
\usepackage[T1]{fontenc}    
\usepackage{url}            
\usepackage{booktabs}       
\usepackage{amsfonts}       
\usepackage{nicefrac}       
\usepackage{microtype}      
\usepackage{xcolor}         

\usepackage{multirow}
\usepackage{makecell}
\usepackage{acronym}
\usepackage[breaklinks,colorlinks,citecolor=blue,linkcolor=red,urlcolor=teal]{hyperref} 
\usepackage[capitalize]{cleveref}
\crefname{section}{Sec.}{Secs.}
\Crefname{section}{Section}{Sections}
\Crefname{table}{Table}{Tables}
\crefname{table}{Tab.}{Tabs.}

\newcommand{\la}{\mathcal{L}_{\mathrm{in}}}
\newcommand{\lb}{\mathcal{L}_{\mathrm{out}}}

\newacro{ViT}[ViT]{Vision Transformer}
\newacro{DINO}[DINO]{Self-Distillation with No Labels}
\newacro{ROI}[ROI]{region of interest}
\newacro{PHS}[PHS]{prompt-guided attention head selection}
\newacro{FOIR}[FOIR]{focus-oriented image retrieval}
\newacro{mIoU}[mIoU]{mean Intersection over Union}
\newacro{Acc}[Acc]{pixel-wise accuracy}
\newacro{mAP}[mAP]{mean Average Precision}
\newacro{CorLoc}[CorLoc]{Correct Localization}

\title{Training Vision Transformers to Focus: Emergence of Spatial Attention Head Specialization}

\author{
  Yuji Nozawa \\
  Kioxia Corporation \\
  \texttt{yuji1.nozawa@kioxia.com} \\
  \And
  Yu-Chieh Lin \\
  Kioxia Corporation \\
  \texttt{yuchieh.lin@kioxia.com} \\
  \And
  Youyang Ng \\
  Kioxia Corporation \\
  \texttt{youyang.ng@kioxia.com} \\
}

\begin{document}
\maketitle

\begin{abstract}
Self-supervised learning in Vision Transformer models, such as DINO, enables robust visual perception without requiring labeled data by leveraging large-scale image datasets. 
However, these models lack a focusing mechanism during training, resulting in perceptual representations that are often ambiguous or overly similar. 
In this work, we introduce two visual focusing loss functions designed to establish correspondence between the model's attention and specific regions within images. 
Specifically, we leverage the multi-head attention mechanism within the model to selectively steer the model's focus, leading to the emergence of spatial attention head specialization, characterized by diverse and spatially focused perceptual capabilities without requiring annotations. 
Through both qualitative and quantitative evaluations, we demonstrate that our method substantially increases the spatial selectivity and diversity of attention heads, improving both interpretability and performance on dense prediction tasks such as zero-shot image segmentation.
\end{abstract}
\begin{figure}[h]
    \centering
    \includegraphics[width=\linewidth]{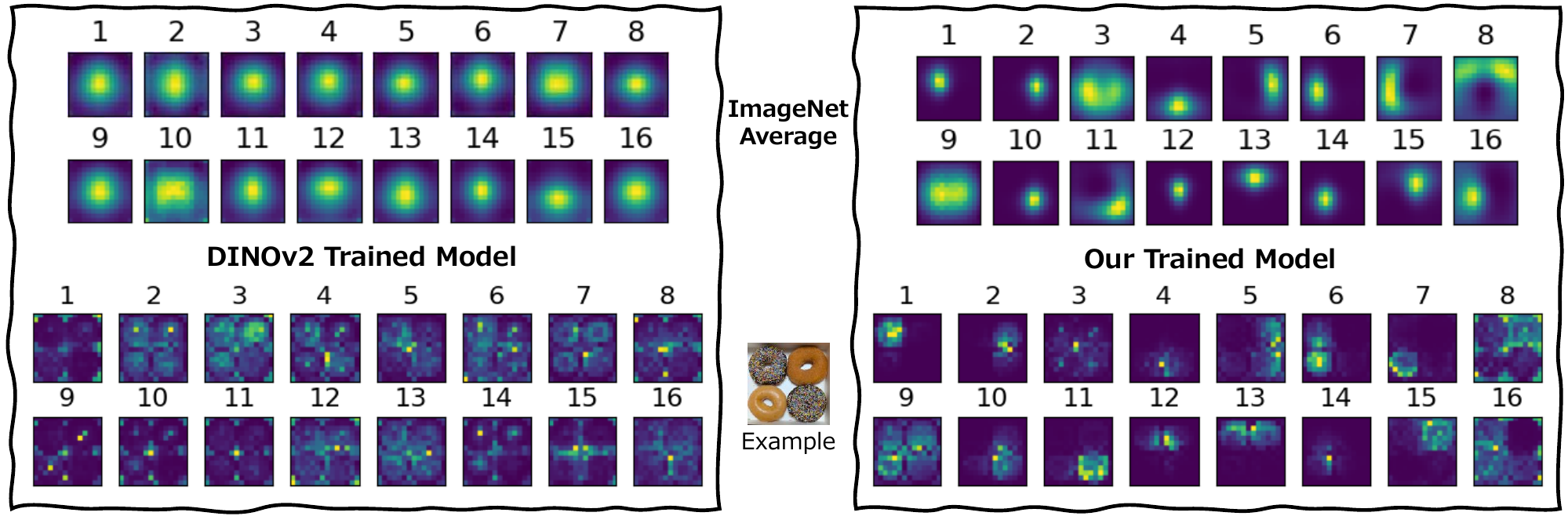}
    \caption{
    Comparison of attention maps produced by the DINOv2 trained model and our trained model. For each model, we show the attention maps averaged over the validation set of the ImageNet-1k~\citep{russakovsky2015imagenet} and the attention maps obtained from a single example image.
    Existing learned perceptual representations (left) in individual attention heads often tend to be ambiguous or overly similar within the example image and across images in the dataset, which limits their interpretability. 
    In contrast, our method (right) selectively steers the focus, leading to the emergence of diverse and spatially focused perceptual capabilities. 
    (Images: Visualization of 16 attention heads in the last layer of trained ViT-L/16-Reg model)}
    \label{fig1}
\end{figure}
\section{Introduction}
\label{sec_intro}
Self-supervised learning has become a powerful paradigm in computer vision, enabling models to acquire rich and meaningful visual representations without the need for costly and time-consuming labeled data. 
Among various architectures, \acp{ViT}~\citep{dosovitskiy2021an} have attracted considerable attention for their remarkable ability to capture long-range dependencies and complex visual patterns within images. 
Methods such as \ac{DINO}~\citep{DINObib} have demonstrated that large-scale unlabeled image datasets can effectively train \acp{ViT} to produce robust and transferable visual features. 
Despite these advances, a fundamental challenge remains: unlike supervised approaches that rely on explicit guidance via labels or self-supervised approaches that incorporate prior knowledge to guide attention, self-supervised \ac{ViT} models without such explicit guidance or prior knowledge typically lack a focusing mechanism during training. 
Consequently, the learned perceptual representations often tend to be ambiguous or overly similar across attention heads, which limits their discriminative capacity and interpretability.

Addressing this limitation is crucial for extending the applicability of self-supervised \acp{ViT}, 
particularly in tasks that require dense representations such as object segmentation and interpretability analysis. 
To this end, we propose two visual focusing loss functions that establish correspondence between the model's attention 
and specific regions within images during training. 
We refer to this correspondence as \emph{region-to-focus correspondence} (ReFoCo). 
The first loss, the \emph{inside-focus loss}, is designed to align the model's focus on local regions within images with the cropped images of these corresponding regions. 
The second loss is inspired by weakly-supervised learning with masked images as proposed in \cite{Singh_2017_ICCV}, 
where masking forces the model to focus not only on the most discriminative object parts, but also on other regions. 
Our proposed second loss, the \emph{outside-focus loss}, is designed to achieve a similar effect in a self-supervised manner by associating partially masked images with the model's focus on regions outside the masked parts. 

Our approach leverages the multi-head attention mechanism in \acp{ViT} to selectively steer the model's focus, 
and we find that diverse and spatially focused perceptual capabilities emerge in our trained model, as shown in \cref{fig1}.
Our method also substantially improves the model's interpretability.

We conduct extensive qualitative and quantitative evaluations to validate the effectiveness of our approach. 
The results demonstrate that incorporating our visual focusing losses substantially improves the model's ability to localize 
and differentiate between various image regions. 
These improvements lead to higher zero-shot image segmentation performance, unsupervised object discovery performance, and multi-object image retrieval performance, 
along with clearer and more interpretable attention maps, highlighting the practical advantages of our method.

We note that while the proposed losses encourage the emergence of diverse and spatially focused perceptual capabilities, 
we observe a slight reduction in classification performance. 
Accordingly, our results should not be interpreted as demonstrating universally stronger global or dense representations. 
Rather, our method aims to introduce focusing mechanisms into self-supervised ViT learning without requiring annotations, 
encouraging attention heads to capture diverse and spatially focused visual information.

Overall, our contributions are as follows:
\begin{itemize}
    \item We identify the ambiguity and similarity of perceptual representations present in current self-supervised ViT models, which limits their interpretability.
    \item We present a new direction for introducing focusing mechanisms into self-supervised \acp{ViT} learning without requiring annotations, using visual focusing loss functions that link representation learning more directly to visual understanding.
    \item We evaluate the performance of models trained using our visual focusing loss functions and demonstrate that they achieve higher zero-shot image segmentation performance, unsupervised object discovery performance, and multi-object image retrieval performance, while also producing more interpretable and meaningful attention maps.
\end{itemize}

\section{Related work}
\label{sec_related}
\paragraph{Vision Transformers and self-supervised learning.}
The advent of \acp{ViT}~\citep{dosovitskiy2021an} has ushered in a paradigm shift in self-supervised learning. 
Leveraging their global receptive fields and adaptive attention mechanisms, 
\acp{ViT} effectively capture long-range dependencies that conventional convolutional neural networks~\citep{simonyan2015very,kaiming2016res} typically struggle to model. 
Techniques such as \ac{DINO}~\citep{DINObib} and iBOT~\citep{zhou2021ibot} have demonstrated that ViTs are effective at learning structured, semantically rich representations without the need for labeled data. 
Remarkably, both \ac{DINO} and its successor, DINOv2~\citep{oquab2023dinov2}, shows that ViTs can generate attention maps closely aligned with object boundaries, suggesting an emergent capacity for dense visual recognition.

\paragraph{Visual focusing mechanisms.}
However, existing self-supervised \ac{ViT} methods often lack effective focusing mechanisms during training. 
Consequently, multiple attention heads tend to converge toward similar activation patterns, producing ambiguous or redundant representations~\citep{Bhojanapalli_2021_ICCV}. 
Some approaches have attempted to mitigate this issue by promoting attention map diversification~\citep{Ren_2025_CVPR}, employing block dropping~\citep{Bhojanapalli_2021_ICCV}, or knowledge distillation~\citep{pmlr-v162-wu22c} techniques to reduce redundancy. 
However, these strategies are typically indirect, relying on architectural heuristics rather than explicitly designed loss functions that encourage interpretability or spatial differentiation. 
Post-hoc explanation methods, such as attention rollout~\citep{abnar_2020_quantifying,Jo_2025_ICIP} and gradient-based techniques~\citep{Selvaraju_2017_ICCV}, 
can reveal where the model focuses after training but do not influence the training dynamics. 
Other efforts have introduced architectural modifications~\citep{Chen_2023_WACV}, add-on modules~\citep{lee2021smallsize}, or prior knowledge-guidance~\citep{miao2022prior} to foster diversity across attention heads or enhance spatial localization. 
Yet these methods often require supervision or additional annotations, limiting their applicability in fully self-supervised settings.

In summary, although current methods successfully capture global structure and high-level semantics, they often lack mechanisms that promote diverse, meaningful, and discriminative attention across different image regions. 
The proposed visual focusing losses offer a principled approach to encourage spatial specialization, enhancing interpretability within self-supervised \ac{ViT} frameworks. 
Integrating such focusing mechanisms that shape attention during training is a relatively unexplored direction with the potential to link self-supervised representation learning to transparent and spatially grounded visual reasoning.

\paragraph{Visual prompt techniques.}
Inspired by prompt engineering in natural language processing~\citep{gpt}, a variety of visual prompts have been studied in computer vision 
to intuitively modify the behavior of models in a visual manner~\citep{segmentanything,finegrain,redcircle,falip,prompttuning,fairvpt,promptself,bahng2022visual,Nozawa_2025_CVPR}. 
In this work, to selectively steer the model's focus, we leverage a visual prompting technique called \ac{PHS}~\citep{Nozawa_2025_CVPR}.
In \ac{PHS}, the visual prompt is given as a set of binary values $R$ assigned to each pixel in the image, where each value indicates whether it is included in the \ac{ROI} or not. 
This has the advantage that the shape of the \ac{ROI} is not restricted. 

\section{Proposed method}
\label{sec_method}
\begin{figure}
    \centering
    \includegraphics[width=\linewidth]{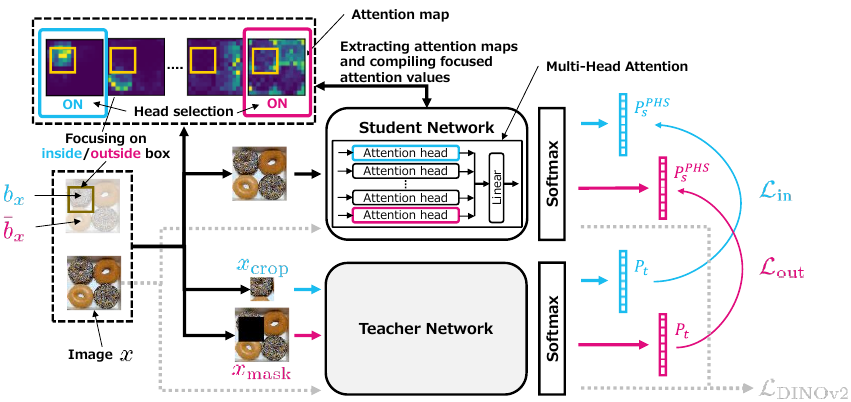}
    \caption{Overview of our proposed visual focusing loss functions $\la$ and $\lb$.}
    \label{fig2}
\end{figure}

We consider self-supervised learning settings, such as \ac{DINO}~\citep{DINObib}, where teacher and student models are used and training is performed by minimizing a loss function between the two models. 
To enable the model to acquire the ability to focus on the \ac{ROI}, we propose to introduce two visual focusing loss functions as shown in \cref{fig2}.

First, we briefly introduce the PHS method~\citep{Nozawa_2025_CVPR} for modifying the model's output using the \ac{ROI}.
Once the \ac{ROI} $R$ is specified, the attention heads that produce the largest outputs of attention maps within the \ac{ROI} are selected.
By applying the head selection method~\citep{gong2021pay,Meng_2022_CVPR,MHSTbib,pmlr-v162-wu22c,nicolicioiu2023learning,basile2025head} that emphasizes the outputs of the selected heads, 
the model's overall output is modified to align with the \ac{ROI}.
Formally, let $H=\{1,2,\dots,h\}$ denote the set of the heads, and let $i_{\mathrm{on}}\subseteq H$ be the set of the selected heads.
The Multi-Head Attention in the last attention layer, $\mathrm{Linear}\left(\left[\mathrm{head}_{1},\mathrm{head}_{2},\dots,\mathrm{head}_{h}\right]\right)$ 
is then modified by replacing $\mathrm{head}_{i}$ with $\mathrm{head}^{\mathrm{HS}}_{i}$, defined by 
\begin{align}
  \mathrm{head}_{i}^{\mathrm{HS}} &= 
      \begin{cases}
        \mathrm{head}_{i} \times h/h_\mathrm{on} & \text{if $i \in i_\mathrm{on}$,} \\
          0 & \text{if $i \notin i_\mathrm{on}$,}
      \end{cases}
\end{align}
where $h_{\mathrm{on}}=|i_{\mathrm{on}}|$ is the number of the selected heads. As a result, for each input image $x$, the model's output $P(x)$ changes to $P^{\mathrm{PHS}}(x, R)$.

\paragraph{Inside-focus loss.}
The first loss function aims to align the model's focus on local areas within images with the cropped images of these corresponding areas. 
For each training image $x$, we designate a box $b_{x}$ within the image as the \ac{ROI}.
Then, the cropped image corresponding to the box, denoted as $x_\mathrm{crop}$, is input to the teacher model.
Meanwhile, the entire image is input to the student model. 
However, instead of simply providing the input to the student model, we apply the \ac{PHS}~\citep{Nozawa_2025_CVPR} method, which enhances the intermediate outputs related to the \ac{ROI} within the model as described above.
The inside-focus loss function $\la$ is defined as the cross-entropy loss between the outputs of the teacher and student networks:
\begin{align}
\label{eq_la}
    \la = - \frac{1}{B}\sum_{x} P_{\mathrm{t}}(x_\mathrm{crop})\log P^{\mathrm{PHS}}_{\mathrm{s}}(x, R=b_{x}),
\end{align}
where $B$ is the batch size, the summation runs over the batch, $P_{\mathrm{t}}$ denotes the softmax output of the teacher network, and $P^{\mathrm{PHS}}_{\mathrm{s}}$ represents the softmax output of the student network with \ac{PHS} visually prompting the box $b_{x}$ applied. 

\paragraph{Outside-focus loss.}
To enable the model to focus not only on the most discriminative object or part but also on other regions, we introduce another loss function. 
Similar to the inside-focus loss function $\la$, we designate a box $b_{x}$ as the \ac{ROI} for each image $x$.
However, for the teacher model in this case, we input the image in which the region within the box is masked, denoted as $x_\mathrm{mask}$. 
For the student model, the entire image is input, similar to the case of $\la$. The difference is the use of \ac{PHS}, which utilizes the external region of the box, $\bar{b}_{x}$, as a visual prompt.
The outside-focus loss function $\lb$ is then given by the following cross-entropy loss:

\begin{align}
\label{eq_lb}
    \lb = - \frac{1}{B}\sum_{x} P_{\mathrm{t}}(x_\mathrm{mask})\log P^{\mathrm{PHS}}_{\mathrm{s}}(x, R=\bar{b}_{x}).
\end{align}

We propose that by combining these visual focusing loss functions with conventional loss functions used in self-supervised learning, the model effectively acquire the ability to focus on various objects or regions within the images. 

\section{Experiments}
\label{sec_experiment}
\subsection{Interpretability comparison of attention maps for conventional training methods}
\label{subsec_attention_training_methods}
\begin{figure}
    \centering
    \includegraphics[width=\linewidth]{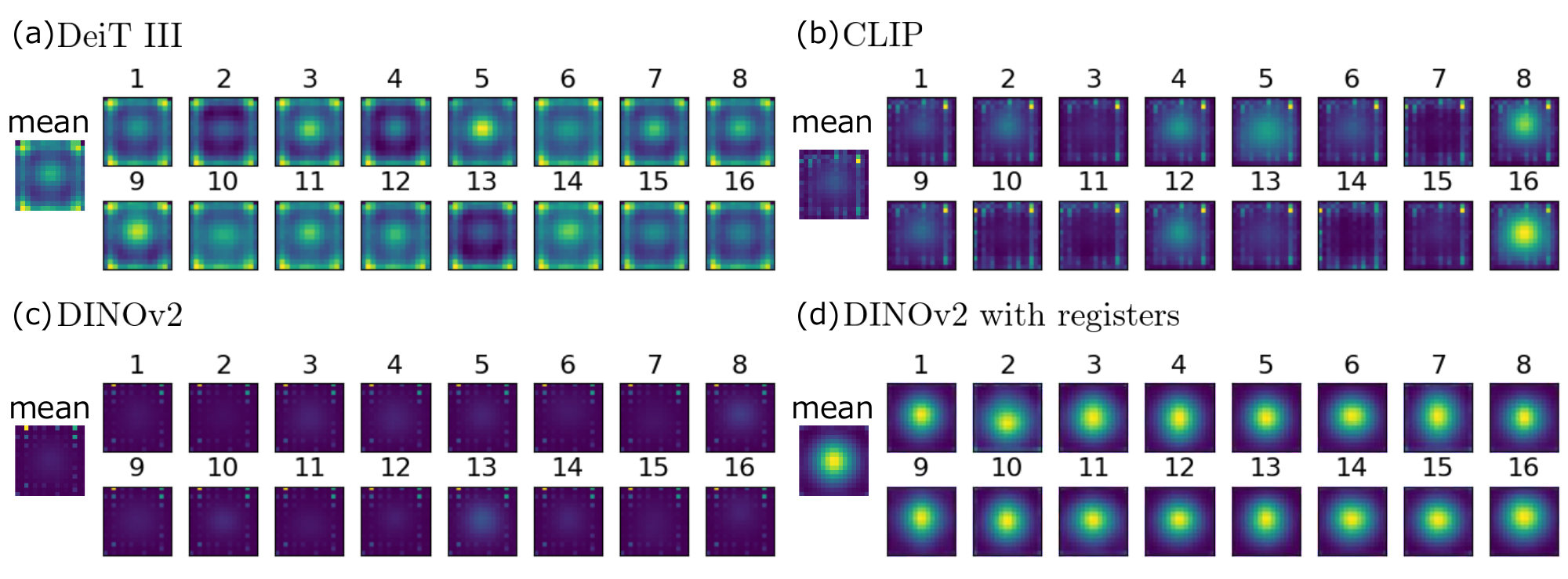}
    \caption{Comparison of the average of attention maps on ImageNet-1k across four training methods using pretrained models (a)-(d).}
    \label{fig_attention_average_imagenetseg_openmodel}
\end{figure}

Since our training method utilizes attention maps via \ac{PHS}, interpretability of the maps is crucial. 
To determine the training method adopted in our method, we compare the attention maps of four different training methods:
DeiT III~\citep{DeiTIII}, CLIP~\citep{clip}, DINOv2~\citep{oquab2023dinov2} and DINOv2 with registers~\citep{darcet2023vision}, utilizing publicly available pretrained models.
DeiT III is a label-supervised training method for \acp{ViT}. 
CLIP is a text-supervised training method using contrastive learning on image-text pairs.
DINOv2 with and without registers are self-supervised training methods to learn robust visual representations. 
DINOv2 with registers extends DINOv2 by introducing additional register tokens to get rid of high-norm tokens in attention maps, which is called artifacts~\citep{darcet2023vision}.
More details about the pretrained models are described in \cref{subsec_training_details}.
\Cref{fig_attention_average_imagenetseg_openmodel} shows the average of the attention maps and the head mean of them on the validation set of ImageNet-1k~\citep{russakovsky2015imagenet}.
Artifacts are observed in the maps produced by DeiT III, CLIP, and DINOv2, which is consistent with \cite{darcet2023vision}, whereas the maps generated by DINOv2 with registers are clearer. 
Based on this result and extended analysis presented in \cref{subsec_extended_training_methods}, we adopt DINOv2 with registers as the baseline training method.

\subsection{Training}
We train \ac{ViT} models with registers~\citep{darcet2023vision} using DINOv2~\citep{oquab2023dinov2} as the baseline training method to study the effects of incorporating our proposed loss functions on the model's performance.
We mainly follow the learning process of DINOv2, but we replace the original DINOv2 loss $\mathcal{L}_{\mathrm{DINOv2}} = \mathcal{L}_{\mathrm{DINO}} + \mathcal{L}_{\mathrm{iBOT}} + 0.1\cdot \mathcal{L}_{\mathrm{koleo}}$ with the following loss function:
\begin{equation}
\label{eq_totalloss}
\begin{aligned}
    \mathcal{L}_{\mathrm{focus}} =&~ (1 - w_{\mathrm{in}} - w_{\mathrm{out}})\cdot \mathcal{L}_{\mathrm{DINO}} + w_{\mathrm{in}}\cdot \la + w_{\mathrm{out}}\cdot \lb\\
    &+ \mathcal{L}_{\mathrm{iBOT}} + 0.1\cdot \mathcal{L}_{\mathrm{koleo}},
\end{aligned}
\end{equation}
where $w_{\mathrm{in}}$ and $w_{\mathrm{out}}$ are the weights of the visual focusing loss functions $\la$ and $\lb$ in \cref{eq_la,eq_lb}, respectively.
We note that the DINO loss $\mathcal{L}_{\mathrm{DINO}}$ is a cross-entropy loss of the same type as $\la$ and $\lb$; therefore, its magnitude is comparable to that of $\la$ and $\lb$. 
$\mathcal{L}_{\mathrm{iBOT}}$ is the iBOT loss~\citep{zhou2021ibot}, $\mathcal{L}_{\mathrm{koleo}}$ is the KoLeo loss~\citep{sablayrolles2018spreading}, 
and for $w_{\mathrm{in}}=0$ and $w_{\mathrm{out}}=0$, \cref{eq_totalloss} reduces to the DINOv2 loss.
For each training image $x$, we designate one box $b_{x}$, which is specified as a random local region within the image, analogous to the local crops used in DINO~\citep{DINObib}.
We share the box $b_{x}$ for the computation of both $\la$ and $\lb$.
Notably, because the boxes can be designated randomly during training, our approach requires only the images as training data and does not necessitate any ground truth data, such as the positional information of objects within the images.
We train the \ac{ViT} models on the train split of ImageNet-1k dataset~\citep{russakovsky2015imagenet} without labels and use the teacher model in all evaluation experiments.
Except for the continual learning discussed in \cref{subsec_continual}, the models are trained from scratch.
Unless otherwise stated, we use ViT-L/16 with registers (ViT-L/16-Reg)~\citep{darcet2023vision} and compare models trained under four different loss weight settings: 
(1) DINOv2 ($w_{\mathrm{in}}=0, w_{\mathrm{out}}=0$), (2) DINOv2 with $\la$ ($w_{\mathrm{in}}=0.1, w_{\mathrm{out}}=0$), (3) DINOv2 with $\lb$ ($w_{\mathrm{in}}=0, w_{\mathrm{out}}=0.1$), and (4) DINOv2 with $\la$ and $\lb$ ($w_{\mathrm{in}}=0.05, w_{\mathrm{out}}=0.05$).
More details about the training conditions are described in \cref{subsec_training_details}.

\subsection{Zero-shot image segmentation}
To validate the ability of \ac{ViT} models to focus on various regions within images, we evaluate their zero-shot image segmentation performance.
We follow a standard protocol~\citep{Chefer_2021_CVPR} for evaluating the effectiveness of their heat maps.
For each heat map, we compute the head mean of the attention maps, where we use the \verb|[CLS]| tokens from each head in the last attention layer as the attention maps.
We then binarize the heat map into a foreground/background segmentation map, where the patch mean of the heat map is used for the threshold value.
We utilize two datasets, ImageNet-Segmentation~\citep{Guillaumin_2014_IJCV} and PASCAL VOC~\citep{everingham2010pascalvoc}, for the evaluation following \cite{Chefer_2021_CVPR}~and~\cite{helbling2025conceptattention}.
ImageNet-Segmentation consists of 4,276 single-label images from the validation set of ImageNet-1k~\citep{russakovsky2015imagenet} with annotated segmentations from 445 categories.
For the evaluation on PASCAL VOC, we utilize PASCAL VOC 2012 benchmark~\citep{everingham2015pascalvoc} and evaluate on a multi-class split (1,449 images).
In our multi-class split experiment, all annotated objects are considered as a single foreground.

\paragraph{Quantitative evaluation results.}
To investigate the impact of changes to the loss function, we quantitatively evaluate zero-shot image segmentation performance using three standard evaluation metrics: \ac{mIoU}, \ac{Acc}, and \ac{mAP}.
\Cref{tab_seg_largereg} provides a comparison of zero-shot image segmentation performance for models trained using four different loss functions.
The results indicate that the performance of the DINOv2 models with either $\la$ alone or $\lb$ alone is comparable to that of the original DINOv2 model.
In contrast, the model with both $\la$ and $\lb$ outperforms all the other models across all the evaluation metrics.

\begin{table}
  \caption{Zero-shot image segmentation performance. The DINOv2 model that integrates both $\la$ and $\lb$ outperforms the original DINOv2 model on both ImageNet-Segmentation and PASCAL VOC. (Model: ViT-L/16-Reg)}
  \label{tab_seg_largereg}
  \centering
  \begin{tabular}{c@{\hspace{2pt}}c@{\hspace{2pt}}c@{\hspace{2pt}}c@{\hspace{2pt}}c@{\hspace{2pt}}c@{\hspace{2pt}}c}
    \toprule
    \multicolumn{1}{c}{} & \multicolumn{3}{c}{ImageNet-Seg.} & \multicolumn{3}{c}{PASCAL VOC}\\
    \cmidrule(lr){2-4}
    \cmidrule(lr){5-7}
    Loss                        & \ac{mIoU}~(\%)$\uparrow$    & \ac{Acc}~(\%)$\uparrow$ & \ac{mAP}~(\%)$\uparrow$ & \ac{mIoU}~(\%)$\uparrow$    & \ac{Acc}~(\%)$\uparrow$ & \ac{mAP}~(\%)$\uparrow$\\
    \midrule
    \multicolumn{1}{l}{DINOv2}             & 63.67 & 80.69 & 85.39 & 58.10 & 75.28 & 84.79\\
    \multicolumn{1}{l}{w/ $\la$}           & 63.93 & 80.61 & 86.03 & 57.82 & 74.76 & 85.00\\
    \multicolumn{1}{l}{w/ $\lb$}           & 63.65 & 80.94 & 84.69 & 58.23 & 75.73 & 84.97\\
    \multicolumn{1}{l}{w/ $\la$ \& $\lb$}  & \textbf{67.29} & \textbf{83.20} & \textbf{86.43} & \textbf{60.68} & \textbf{77.36} & \textbf{85.27}\\
    \bottomrule
  \end{tabular}
\end{table}

\subsection{Qualitative analysis and emergent abilities}
\label{subsec_attention_average}
\begin{figure}
    \centering
    \includegraphics[width=\linewidth]{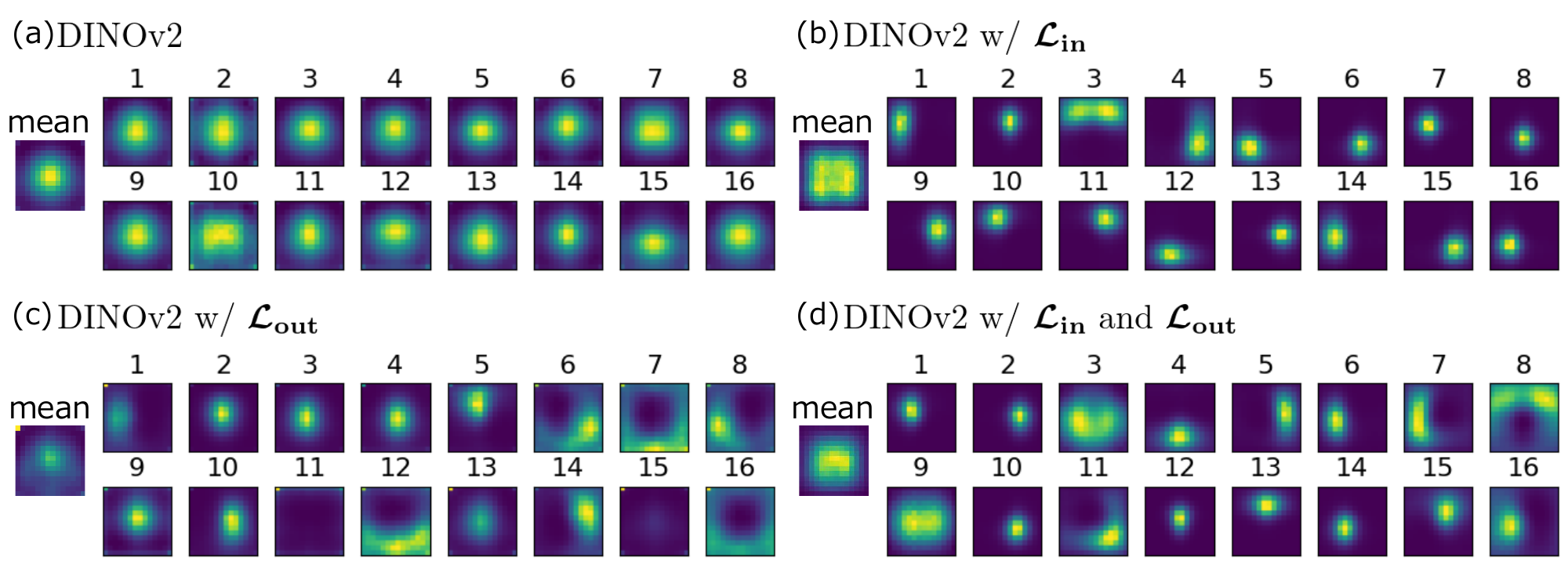}
    \caption{Comparison of the average of attention maps on ImageNet-1k from head$1$ to head$16$ and their head mean. 
    Each model is trained using one of four loss functions (a)-(d).}
    \label{fig_attention_average_imagenetseg_largereg}
\end{figure}

\paragraph{Dataset-scale attention heads visualization.}
To investigate which regions of the images the models focus on, we visualize the average of the attention maps across the entire set of evaluation dataset images.
\Cref{fig_attention_average_imagenetseg_largereg} shows the average of the attention maps and the head mean of them used as the heat map on the validation set of ImageNet-1k.
As a result, we find that the visual loss functions qualitatively change the characteristics of attention maps.
\paragraph{Emergent focusing ability: Spatial Attention Head Specialization.}
As shown in \cref{fig_attention_average_imagenetseg_largereg}(a), the attention maps of each head in the original DINOv2 model mainly concentrate their outputs around the center.
In contrast, as illustrated in \cref{fig_attention_average_imagenetseg_largereg}(b), the model with $\la$ shows substantial variety in the output locations across different heads. 
This visualizes the emergence of the ability for each head to focus on distinct spatial positions, and their information is integrated to produce the comprehensive heat map.
Notably, the focus area of the heat map in this model is broader compared to that of the original model. 
In the model with $\lb$, in addition to heads that primarily focus on the central region as seen in the original model, 
there also emerge heads that tend to avoid the center and focus on other regions. 
This behavior is considered to reflect the design of $\lb$, which encourages attention to objects other than the most discriminative objects.
Additionally, a large activation is observed in a single patch at the top-left corner, which is considered to be a type of artifact, as discussed in \cite{darcet2023vision}.
As described above, by using $\la$ and $\lb$, various attention behaviors emerge across different heads. 
In the model with both $\la$ and $\lb$, we observe the coexistence of the attention maps observed in the model with $\la$ alone or $\lb$ alone, in addition to attention maps that focus on broader regions. 
We note that the artifacts seen in the model with $\lb$ alone are absent.
While standard DINOv2 already exhibits content-based head specialization, with individual heads capturing different semantic features within an image, 
our method retains such content-based specialization while introducing an additional, complementary form of specialization across spatial regions. 
We refer to this phenomenon as the emergence of spatial attention head specialization.

\subsection{Empirical analysis of segmentation capability}
\label{sebsec_attention}
\begin{figure}
    \centering
    \includegraphics[width=\linewidth]{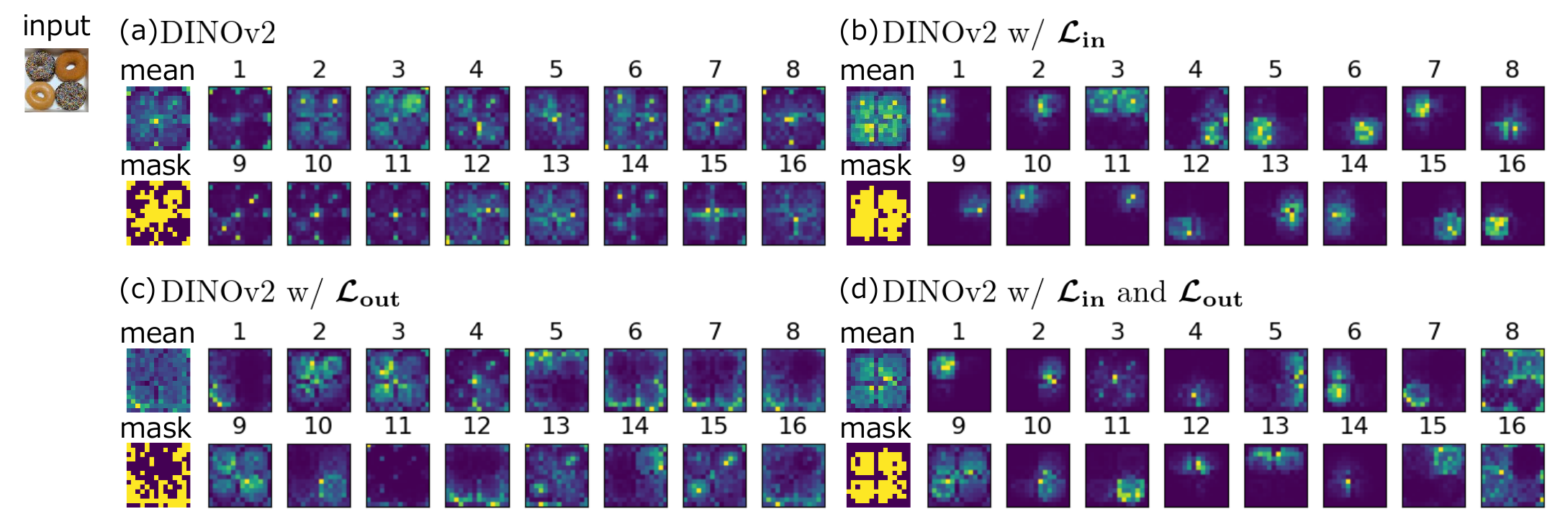}
    \caption{An example of attention maps for each model (a)-(d). The image in the top left is the ``input''. 
    The images labeled ``mean'' represent the head means of the attention maps used as heat maps, and the images labeled ``mask'' represent the segmentation masks.}
    \label{fig_attention_example}
\end{figure}

To understand how segmentation works on individual images, we analyze examples of attention maps for a specific input image, as shown in \cref{fig_attention_example}. 
In the case of the original DINOv2 model, we observe that all heads focus at least partially on the box behind the donuts, which leads to ambiguous segmentation. 
We speculate that the original model pays close attention to the box because part of the box is visible in the center of the input image. 
When $\la$ is introduced, we observe that the model focus on object regions corresponding to the parts discussed in the attention map averages in \cref{fig_attention_average_imagenetseg_largereg}. 
For instance, head1, 7, 11, and 15 in \cref{fig_attention_example}(d) are able to focus primarily on each donut region in the image. 
Furthermore, when $\lb$ is introduced, as seen in head8 of \cref{fig_attention_example}(d), there emerge attention heads that focus on the box regions while avoiding the donuts area. 
This corresponds to the heads focusing on non-central regions on average, as discussed in \cref{subsec_attention_average}.

\subsection{Segmentation performance by position}
\begin{figure}
    \centering
    \includegraphics[width=\linewidth]{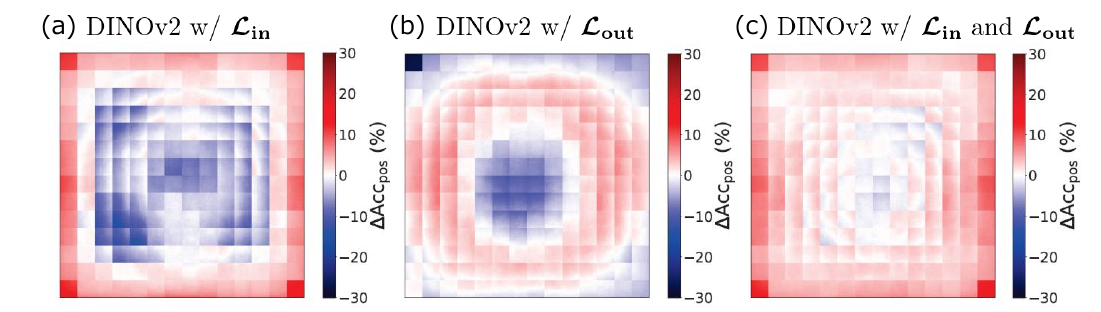}
    \caption{Changes in segmentation accuracy at each position compared to the accuracy of the original DINOv2 model.}
    \label{fig_acc_pos_imagenet_largeref}
\end{figure}

As shown in \cref{tab_seg_largereg}, we find that the model with $\la$ and $\lb$ exhibits higher segmentation performance compared to the other models evaluated.
To investigate the positional dependence of performance among the models, we calculate the accuracy at each position for each model and examine the changes in accuracy from the original model, referred to as $\Delta \mathrm{Acc}_{\mathrm{pos}}$ as shown in \cref{fig_acc_pos_imagenet_largeref}.
The figure shows that $\la$ and $\lb$ play complementary roles in improving segmentation accuracy. 
First, as shown in \cref{fig_acc_pos_imagenet_largeref}(a), the model with $\la$ alone decreases accuracy in the central region while improving accuracy in the peripheral regions.
This result suggests that whereas the original DINOv2 model tends to focus on objects near the center, $\la$ encourages attention to various regions, which enhances segmentation performance in less salient peripheral areas while somewhat reducing focus on the center.
In contrast, \cref{fig_acc_pos_imagenet_largeref}(b) shows that the model with $\lb$ alone exhibits improved accuracy in a ring-shaped region surrounding the center. 
This result can be interpreted as indicating that $\lb$ primarily enhances segmentation performance in regions surrounding central objects, which are typically located in the center.
Although each of these two models shows improved segmentation performance in certain regions, the decline in accuracy around the central area results in overall performance comparable to the original DINOv2 model. 
Finally, \cref{fig_acc_pos_imagenet_largeref}(c) shows that the model with both $\la$ and $\lb$ successfully integrates the advantages of both loss functions. 
Furthermore, the accuracy drop observed in the central region in the models (a) and (b) is substantially mitigated in the model (c), resulting in an improvement in \ac{Acc}.

\subsection{Unsupervised object discovery}
The ability to focus on a wide range of areas within images is also considered effective for object discovery. Therefore, we evaluate the unsupervised object discovery performance of our trained models.
For the evaluation, we use LOST~\citep{LOST}, which infers bounding boxes for objects from the patch tokens of features in the attention layers.
Following \cite{darcet2023vision}~and~\cite{jiang2025vision}, we present the best results of \ac{CorLoc} across the key, query, or value features over the last four attention layers, and details of the evaluation setup are described in \cref{subsec_evaluation_details}.
We utilize three datasets, VOC07 trainval, VOC12 trainval~\citep{everingham2010pascalvoc}, and COCO 20k~\citep{lin2014coco,Vo20rOSD}.
\Cref{tab_discovery_largereg} presents the performance of four trained models on the unsupervised object discovery task. 
We observe that using $\la$ or $\lb$ alone results in lower performance than the original DINOv2 model, except in the case of $\la$ on COCO. 
However, combining $\la$ and $\lb$ outperforms the original model in all the three datasets.
This improvement through combination is consistent with the results in the zero-shot image segmentation task. 

\begin{table}
  \centering
  \begin{minipage}{0.50\linewidth}
    \centering
    \caption{Unsupervised object discovery performance. (Model: ViT-L/16-Reg)}
    \label{tab_discovery_largereg}
    \begin{tabular}{cccc}
      \toprule
      Loss                                   & VOC07 & VOC12 & COCO\\
      \midrule
      \multicolumn{1}{l}{DINOv2}             & 58.6 & 63.3 & 45.4 \\
      \multicolumn{1}{l}{w/ $\la$}           & 56.0 & 59.5 & \textbf{45.7} \\
      \multicolumn{1}{l}{w/ $\lb$}           & 42.0 & 47.3 & 35.1 \\
      \multicolumn{1}{l}{w/ $\la$ \& $\lb$}  & \textbf{61.6} & \textbf{64.6} & 45.5 \\
      \bottomrule
    \end{tabular}
  \end{minipage}
  \hfill
  \begin{minipage}{0.45\linewidth}
    \caption{ImageNet-1k classification performance. (Model: ViT-L/16-Reg)}
    \label{tab_classification_largereg}
    \begin{tabular}{ccc}
        \toprule
        Loss & k-NN~(\%) & Linear~(\%) \\
        \midrule
        \multicolumn{1}{l}{DINOv2}            & \textbf{81.7} & \textbf{82.6} \\
        \multicolumn{1}{l}{w/ $\la$}          & 79.1 & 81.4 \\
        \multicolumn{1}{l}{w/ $\lb$}          & 79.7 & 81.7 \\
        \multicolumn{1}{l}{w/ $\la$ \& $\lb$} & 80.3 & 81.8 \\
        \bottomrule
    \end{tabular}
  \end{minipage}
\end{table}

\subsection{Performance on other tasks}
In addition to zero-shot image segmentation and unsupervised object discovery, we conduct additional experiments on two complementary tasks: a classification task and an image retrieval task. 
\paragraph{Classification performance.}
We evaluate our models on the ImageNet-1k classification task with k-NN ($k=20$) and linear probing.
\Cref{tab_classification_largereg} presents the top-1 accuracy on the validation set of ImageNet-1k.
For k-NN classification, we observe that the accuracy drops by approximately $1-3\%$ when visual focusing loss functions are introduced compared to the original DINOv2 model.
Our possible explanation for the performance drop is as follows. 
The classification task is formulated as a single-label problem, where images typically contain a dominant object near the center.
This task setting aligns with local-to-global correspondence for single objects in \ac{DINO}~\citep{DINObib}.
In contrast, our proposed loss functions are designed to capture a wide range of regions and focus on various locations, and may therefore be less specialized for this setting.
For linear probing classification, the accuracy difference between our visual focusing models and the original DINOv2 model is smaller than that observed with k-NN classification. 
This result suggests that training a linear classifier for classification purposes enables effective utilization of the frozen features that have been pretrained to focus.

\paragraph{Image retrieval performance.}
We also evaluate our models using an image retrieval task called \ac{FOIR}~\citep{Nozawa_2025_CVPR}.
\Ac{FOIR} task is a multi-object image retrieval task that evaluates whether models can retrieve appropriate images for each specified object in query images, where the query images may contain multiple objects.
We utilize COCO~\citep{lin2014coco} and PASCAL VOC~\citep{everingham2010pascalvoc} datasets for evaluation.
We use Mean Precision at $k$ ($\mathrm{MP@}k$) as an evaluation metric to measure the relevance of the top-$k$ retrieved results.
\Cref{tab_foir_largereg} presents $\mathrm{MP@}100$ for each model and dataset.
We evaluate models using two retrieval schemes. 
The first scheme, referred to as ``Fixed per image'', utilizes a single retrieval result obtained by extracting one feature vector from each query image for image retrieval. 
The second scheme, ``Visual prompt per object'', applies \ac{PHS} during inference and specifies a bounding box for each object as a visual prompt, thereby obtaining retrieval results that focus on individual objects. 
First, the consistent performance gain of the visual prompt per object scheme over the fixed per image one across the models indicates that \ac{PHS} is effective during inference. 
In a comparison between the models, under the fixed per image scheme, the original DINOv2 model outperforms the model with $\la$ and $\lb$; 
however, under the visual prompt per object scheme, the trend reverses, with the model with $\la$ and $\lb$ outperforming the original DINOv2 model. 
This result suggests that the feature representations of DINOv2 effectively capture global image characteristics, resulting in averaged retrieval performance over objects, 
while the model with $\la$ and $\lb$, consistent with its training objective, exhibits its strengths when focusing on individual objects.

\begin{table}
    \caption{Focus-Oriented Image Retrieval performance. (Model: ViT-L/16-Reg)}
    \label{tab_foir_largereg}
    \centering
    \begin{tabular}{c@{\hspace{2pt}}c@{\hspace{2pt}}c@{\hspace{2pt}}c}
        \toprule
         &  & COCO & PASCAL VOC\\
        \cmidrule(lr){3-3}
        \cmidrule(lr){4-4}
        Loss & Retrieval scheme & $\mathrm{MP@}100~(\mathrm{\%})\uparrow$ & $\mathrm{MP@}100~(\mathrm{\%})\uparrow$\\
        \midrule
        \multicolumn{1}{l}{DINOv2}              & \multicolumn{1}{l}{Fixed per image}          & 54.2 & 72.8\\
        \multicolumn{1}{l}{DINOv2}              & \multicolumn{1}{l}{Visual prompt per object} & 55.2 & 74.2\\
        \multicolumn{1}{l}{w/ $\la$ \& $\lb$}   & \multicolumn{1}{l}{Fixed per image}          & 52.7 & 72.4\\
        \multicolumn{1}{l}{w/ $\la$ \& $\lb$}   & \multicolumn{1}{l}{Visual prompt per object} & \textbf{55.9} & \textbf{75.8}\\
        \bottomrule
    \end{tabular}
\end{table}

\subsection{Additional results on model size variations}
\label{subsec_model_size}
To evaluate the robustness of our method, we further examine its effectiveness across different model sizes. 
\Cref{tab_seg_baseandgiantreg} presents the zero-shot image segmentation performance of the original DINOv2 model and the model with $\la$ and $\lb$ for ViT-B/16-Reg and ViT-g/16-Reg, 
evaluated in the same manner as for ViT-L/16-Reg. 
These results demonstrate that our method consistently improves performance across all evaluated model sizes, confirming its robustness. 

\begin{table}
  \caption{Zero-shot image segmentation performance. (Model: ViT-\{Size\}/16-Reg)}
  \label{tab_seg_baseandgiantreg}
  \centering
  \begin{tabular}{c@{\hspace{2pt}}c@{\hspace{2pt}}c@{\hspace{2pt}}c@{\hspace{2pt}}c@{\hspace{2pt}}c@{\hspace{2pt}}c@{\hspace{2pt}}c}
    \toprule
    \multicolumn{2}{c}{} & \multicolumn{3}{c}{ImageNet-Seg.} & \multicolumn{3}{c}{PASCAL VOC}\\
    \cmidrule(lr){3-5}
    \cmidrule(lr){6-8}
    Size & Loss                        & \ac{mIoU}~(\%)$\uparrow$    & \ac{Acc}~(\%)$\uparrow$ & \ac{mAP}~(\%)$\uparrow$ & \ac{mIoU}~(\%)$\uparrow$    & \ac{Acc}~(\%)$\uparrow$ & \ac{mAP}~(\%)$\uparrow$\\
    \midrule
    \multirow{2}{*}{B}  &\multicolumn{1}{l}{DINOv2}            & 68.32 & 83.81 & 87.63 & 62.20 & 78.14 & 86.46\\
                        &\multicolumn{1}{l}{w/ $\la$ \& $\lb$} & \textbf{69.33} & \textbf{84.40} & \textbf{87.84} & \textbf{62.98} & \textbf{78.65} & \textbf{86.77}\\
    \midrule
    \multirow{2}{*}{g}  &\multicolumn{1}{l}{DINOv2}            & 63.37 & 80.53 & 85.20 & 58.34 & 75.55 & 84.85\\
                        &\multicolumn{1}{l}{w/ $\la$ \& $\lb$} & \textbf{67.61} & \textbf{83.31} & \textbf{86.86} & \textbf{61.31} & \textbf{77.65} & \textbf{86.09}\\
    \bottomrule
  \end{tabular}
\end{table}

\section{Discussion}
\label{sec_discussion}
\paragraph{Local-to-global and region-to-focus correspondence.}
Our results highlight a distinction between conventional local-to-global correspondence~\citep{DINObib} and 
region-to-focus correspondence introduced by our visual focusing losses. 
The main DINOv2 objective is based on local-to-global correspondence and is well suited to images containing a large dominant object, 
as it encourages correspondence between the global representation of the object and local representations of its parts.
In contrast, our method is designed for images containing multiple objects or semantic regions, a common setting in natural images. 
To address this setting, in addition to establishing local-to-global correspondence, 
our proposed loss functions encourage correspondence between regional representations obtained through masking or cropping and selective attention toward the corresponding regions. 
As observed in our experiments, this encourages the emergence of spatial attention head specialization, 
with performance gains in zero-shot image segmentation, unsupervised object discovery, and multi-object image retrieval. 

\paragraph{Practical implications.}
These properties may also have practical implications for interpretability and customizability. 
Distinct and spatially specialized attention heads can provide a more structured view of the visual information 
captured by the model, potentially facilitating the analysis of its behavior and failure cases. 
Moreover, such specialization may offer a straightforward mechanism for feature manipulation during inference. 
For example, the output features could be modified by dynamically disabling the signals from individual attention heads during inference, 
thereby potentially providing a mechanism for on-the-fly model customization.
These possibilities could be useful in applications that require selective or dynamically varying visual focus, 
such as robotic vision and autonomous driving. 
However, we do not evaluate such applications in this work, 
and their effectiveness remains an important direction for future investigation. 

\section{Conclusion}
\label{sec_conclusion}
Our work addresses a key limitation of self-supervised \ac{ViT} models 
by introducing visual focusing loss functions that establish correspondence between
the model's attention and specific regions within images. 
We find that diverse and spatially focused perceptual capabilities emerge in our trained model, 
enabling the model to leverage different attention heads to focus on distinct and meaningful image regions. 
This targeted focusing mechanism effectively mitigates the ambiguity and redundancy often observed in learned representations, thereby improving the interpretability of the model's attention. 
Through comprehensive experiments, we demonstrate that our approach improves performance in zero-shot image segmentation, unsupervised object discovery, and multi-object image retrieval, while also producing clearer and more informative attention maps that support better visual understanding. 
Overall, these results highlight the potential of introducing focusing mechanisms into self-supervised representation learning without requiring annotations, providing a new direction for learning diverse and spatially focused visual information.

{
    \small
    \bibliographystyle{ieeenat_fullname}
    \bibliography{mainbib}
}


\clearpage

\setcounter{equation}{0}
\renewcommand{\theequation}{A\arabic{equation}}
\setcounter{figure}{0}
\renewcommand{\thefigure}{A\arabic{figure}}
\setcounter{table}{0}
\renewcommand{\thetable}{A\arabic{table}}
\setcounter{section}{0}
\renewcommand{\thesection}{A\arabic{section}}

\appendix

\section{Limitations}
While our method can train the model to focus on diverse regions within an image, 
it does not allow explicit control over where each attention head focuses, 
as the regions of focus are determined implicitly during training. 
As a result, it is not possible to know in advance which regions each attention head will focus on until after training. 
Our method also increases the training time and introduces a trade-off between classification performance and performance improvements in 
zero-shot segmentation, object discovery, and \ac{FOIR} tasks.
Exploring other effective ways to utilize the visual focusing image representations 
obtained by our method remains an important direction for future work.

\section{Societal impacts}
In this study, we propose a self-supervised training method, and our trained models demonstrate high-performance segmentation across a wide range of areas within images. 
Our method can identify both salient and subtle objects in images, and it has potential applications such as medical imaging, autonomous driving, and defect detection. 
However, incorrect segmentation of subtle structures may lead to erroneous decisions, posing risks especially in scenarios where safety is critical. 
The ability to extract subtle visible information also has a risk of misuse, such as applying the method to privacy-invasive surveillance. 
To mitigate these concerns, we emphasize clear communication of limitations and intended use.

\section{Details of training and evaluation setups}
\subsection{Training}
\label{subsec_training_details}
Most of the training hyperparameters are identical to those used in \cite{oquab2023dinov2} for training ViT-L/16 on ImageNet-1k. 
Here, we describe the differences from that setting. 
Due to computational resource constraints, we train the models using 4 NVIDIA H200 GPUs (141GB of memory), instead of 32 NVIDIA A100 GPUs (80GB of memory).
Correspondingly, the batch size per GPU is set to 128 for the base and large models, and 64 for the giant models.
In addition, for the models with registers, the number of registers is set to $4$.

\paragraph{Data augmentation.}
For each training image, following DINOv2, we apply multi-crop data augmentation, where global and local crops are generated using random resized cropping, 
together with standard augmentations such as horizontal flipping, color jittering, Gaussian blur, and solarization. 
The global crops are resized to $224 \times 224$ with $\mathrm{scale}=(0.32, 1.0)$ and $\mathrm{ratio}=(3.0 / 4.0, 4.0 / 3.0)$, 
while the local crops are resized to $96 \times 96$ with $\mathrm{scale}=(0.05, 0.32)$ and the same aspect-ratio range. 
When our loss functions $\la$ or $\lb$ are introduced, a box $b_{x}$ for the PHS method is obtained as a region sampled via random resized cropping from the image, 
using the same setting as the local crops. We set the number of selected heads $h_{\mathrm{on}}$ during training to $5$ in our main experiments as discussed in \cref{subsec_numofheads}. 
For $\lb$, we also introduce randomness by selecting either the top or bottom edge of the visual-prompt box with equal probability for each batch and aligning the selected edge at the same height across all images in the batch. 

\paragraph{Training from scratch.} 
The number of training iterations is set to 500,000.
We set the base learning rate, which is the learning rate for a batch size of 1024, to $0.004$ for the base and large models, and $0.002$ for the giant models for stable training.
The training time for ViT-L/16-Reg is approximately 2 days and 22 hours for the DINOv2 model, 3 days and 21 hours for the model with $\la$, 4 days and 11 hours for the model with $\lb$, and 5 days and 10 hours for the model with $\la$ and $\lb$.

\paragraph{Pretrained models and continual learning.}
For our experiments about pretrained models in \cref{subsec_attention_training_methods}~and~\cref{subsec_extended_training_methods}, we compare DeiT III~\citep{DeiTIII}, CLIP~\citep{clip}, DINOv2~\citep{oquab2023dinov2} and DINOv2 with registers~\citep{darcet2023vision}, utilizing publicly available pretrained models.
For DeiT III, we use the ViT-L/16 pretrained on ImageNet-1k at a resolution of $224 \times 224$. 
For CLIP, we use the OpenCLIP implementation~\citep{openclip} and employ the ViT-L/14 image encoder pretrained by OpenAI~\citep{clip}.
For DINOv2 and DINOv2 with registers, we use the ViT-L/14 and the ViT-L/14-Reg, respectively, both pretrained on the large-scale LVD-142M dataset. 
We discuss continual learning using pretrained models in \cref{subsec_continual}. 
To match the patch size of 14, we change the local crops size from 96 to 98.
During the initial training stage of the DINO head only, the number of iterations is set to 50,000.
In the subsequent stage, where the entire networks are trained, we also set the number of iterations to 50,000.
Throughout training, the base learning rate is set to 0.0004, and the number of warmup epochs is set to 1.
The training dataset is the train split of ImageNet-1k dataset~\citep{russakovsky2015imagenet} as in training from scratch.
The training time for the DINO head of ViT-L/14-Reg is approximately 11 hours, and 
the training time for the entire networks is approximately 9 hours for the DINOv2 model, 12 hours for the model with $\la$, 13 hours for the model with $\lb$, 16 hours for the model with $\la$ and $\lb$.

\subsection{Evaluation}
\label{subsec_evaluation_details}
In all our evaluation experiments except unsupervised object discovery, the input images are resized to $224 \times 224$, whereas the unsupervised object discovery task does not involve resizing.

For the unsupervised object discovery task, we utilize three datasets, VOC07 trainval (5,011 images), VOC12 trainval (11,540 images)~\citep{everingham2010pascalvoc}, and COCO 20k (19,817 images)~\citep{lin2014coco,Vo20rOSD}.
Following \cite{darcet2023vision}~and~\cite{jiang2025vision}, we present the best results of \ac{CorLoc} across the key, query, or value features over the last four attention layers, 
sweeping over a manual bias term to the Gram matrix of features. We normalize the features and sweep the bias term over the range $[-1, 0]$ with a step size of $0.01$.

The evaluation procedure for k-NN and linear probing classification follows that of \cite{oquab2023dinov2}.
For linear probing classification, we report the highest accuracy performing the following grid search of three parameters: 
learning rate in \{0.0001, 0.0002, 0.0005, 0.001, 0.002, 0.005, 0.01, 0.02, 0.05, 0.1, 0.2, 0.3, 0.5\}, 
number of output layers in \{1, 4\}, and 
concatenate average-pooled tokens in \{yes, no\}. 
The linear classifier is trained with SGD for 12,500 iterations.

For the \ac{FOIR} task, the evaluation procedure follows that of \cite{Nozawa_2025_CVPR}. 
When the retrieval scheme is Visual prompt per object, the number of selected heads at evaluation is set to $2$.
We use two datasets for this evaluation: COCO~\citep{lin2014coco} and PASCAL VOC~\citep{everingham2010pascalvoc}. 
COCO dataset contains 80 object categories, and we use val2017 for query images (5,000 images) and train2017 for database images (118,287 images).
PASCAL VOC dataset contains 20 object categories, and we use test2007 for query images (4,952 images) and trainval2007+2012 for database images (16,551 images).

\section{Extended analyses}
\subsection{Extended results on interpretability of attention maps for conventional training methods}
\label{subsec_extended_training_methods}
In addition to our experiment in \cref{subsec_attention_training_methods}, we conduct extended analyses on the interpretability of attention maps obtained using four different training methods: 
DeiT III~\citep{DeiTIII}, CLIP~\citep{clip}, DINOv2~\citep{oquab2023dinov2} and DINOv2 with registers~\citep{darcet2023vision}, utilizing publicly available pretrained models.

\begin{figure}
    \centering
    \includegraphics[width=\linewidth]{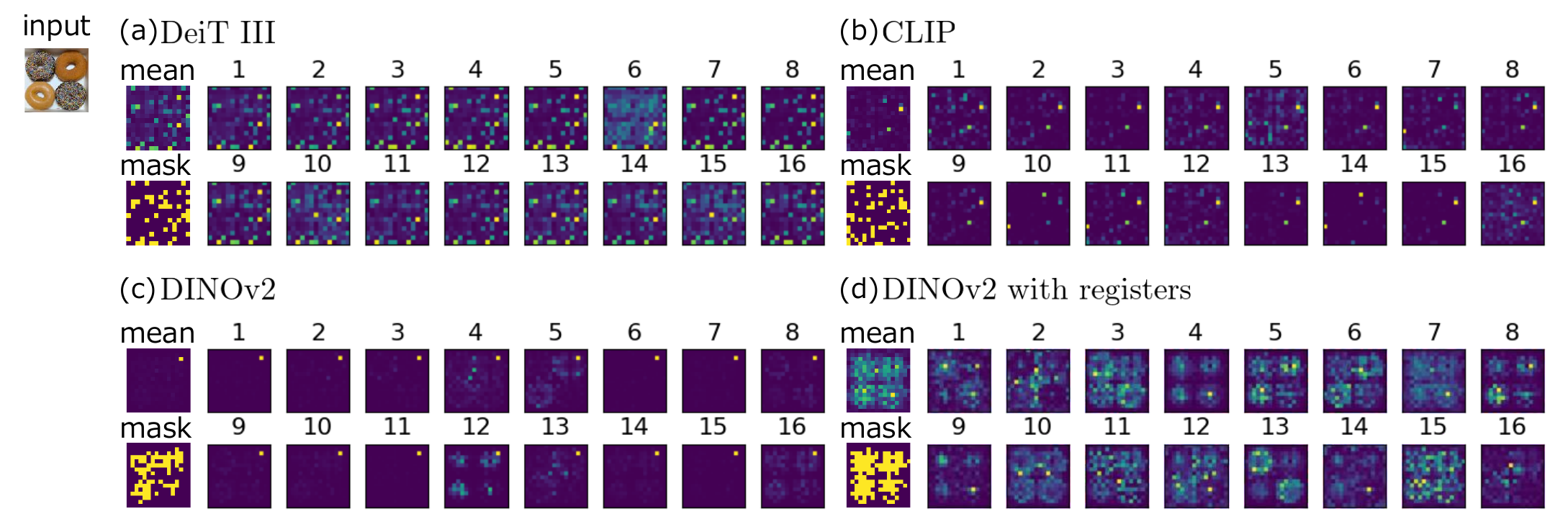}
    \caption{Examples of attention maps for four training methods (a)-(d).}
    \label{fig_attention_example_openmodel}
\end{figure}

\Cref{tab_seg_openmodel} presents a comparison of zero-shot image segmentation performance of each training method. 
This result shows that DINOv2 and DINOv2 with registers outperform the other methods.
As shown in \cref{fig_attention_example_openmodel}, the impact of these artifacts is substantial for DeiT III and CLIP, leading to low interpretability of the attention maps. 
In contrast, DINOv2 produces a single patch that appears to be an artifact, while the remaining patches capture meaningful image information, resulting in a meaningful mask.
In DINOv2 with registers, the register tokens are effective in suppressing artifacts, and the object regions are captured more clearly.

\begin{table}
  \caption{Zero-shot image segmentation performance comparison of four training methods (a)-(d).}
  \label{tab_seg_openmodel}
  \centering
  \begin{tabular}{c@{\hspace{2pt}}c@{\hspace{2pt}}c@{\hspace{2pt}}c@{\hspace{2pt}}c@{\hspace{2pt}}c@{\hspace{2pt}}c}
    \toprule
    \multicolumn{1}{c}{} & \multicolumn{3}{c}{ImageNet-Seg.} & \multicolumn{3}{c}{PASCAL VOC}\\
    \cmidrule(lr){2-4}
    \cmidrule(lr){5-7}
    Method                        & \ac{mIoU}~(\%)$\uparrow$    & \ac{Acc}~(\%)$\uparrow$ & \ac{mAP}~(\%)$\uparrow$ & \ac{mIoU}~(\%)$\uparrow$    & \ac{Acc}~(\%)$\uparrow$ & \ac{mAP}~(\%)$\uparrow$\\
    \midrule
    \multicolumn{1}{l}{\makecell[l]{DeiT III \\ \citep{DeiTIII}}}                  & 38.21 & 59.73 & 74.36 & 37.98 & 58.43 & 77.54\\
    \midrule
    \multicolumn{1}{l}{\makecell[l]{CLIP \\ \citep{clip}}}                         & 47.06 & 72.06 & 76.41 & 45.37 & 69.59 & 77.66\\
    \midrule
    \multicolumn{1}{l}{\makecell[l]{DINOv2 w/o reg. \\ \citep{oquab2023dinov2}}}   & 52.35 & 76.14 & 82.29 & 52.75 & 74.40 & 83.88\\
    \midrule
    \multicolumn{1}{l}{\makecell[l]{DINOv2 w/~~ reg. \\ \citep{darcet2023vision}}} & 53.71 & 73.95 & 81.61 & 55.50 & 74.53 & 84.03\\
    \bottomrule
  \end{tabular}
\end{table}

\subsection{Continual learning in existing models}
\label{subsec_continual}
In this subsection, we consider applying our training method to continual learning in existing models.
For the pretrained model, we utilize the publicly available backbone ViT-L/14-Reg weights pretrained by DINOv2 with registers~\citep{darcet2023vision}, which we evaluate in \cref{subsec_attention_training_methods}~and~\cref{subsec_extended_training_methods}.
Since pretrained weights of the DINO head is not available, we first train the DINO head using DINOv2 with registers from scratch, while keeping the pretrained backbone frozen.
After training the DINO head, we then perform training the entire networks, using each loss function with fewer iterations than training from scratch. 
We use ImageNet-1k as the training dataset, as in the main experiments. 
Details of the training setup are described in \cref{subsec_training_details}.

\Cref{tab_seg_largereg_continual} presents a comparison of zero-shot image segmentation performance between the pretrained model and models further trained using four different loss functions. 
On ImageNet-Segmentation dataset, the performance improves with all loss functions used for continual learning, with the DINOv2 model with $\la$ and $\lb$ achieving the highest performance. 
On PASCAL VOC dataset, the DINOv2 loss, DINOv2 loss with $\la$, and DINOv2 loss with $\lb$ all show performance degradation across all metrics. 
The discrepancy between the results on ImageNet-Segmentation and PASCAL VOC can be attributed to the domain gap between these datasets. 
While further training on ImageNet-1k improves performance on the in-domain dataset ImageNet-Segmentation, it results in degradation in the out-of-domain dataset PASCAL VOC. 
Notably, the DINOv2 loss with both $\la$ and $\lb$ improves performance on two out of three metrics, mIoU and mAP. 
This indicates that combination of $\la$ and $\lb$ exhibits stronger robustness to domain shifts and better generalization ability. 

\begin{table}
  \caption{Zero-shot image segmentation performance in continual learning setting. (Model: ViT-L/14-Reg)}
  \label{tab_seg_largereg_continual}
  \centering
  \begin{tabular}{c@{\hspace{2pt}}c@{\hspace{2pt}}c@{\hspace{2pt}}c@{\hspace{2pt}}c@{\hspace{2pt}}c@{\hspace{2pt}}c}
    \toprule
    \multicolumn{1}{c}{} & \multicolumn{3}{c}{ImageNet-Seg.} & \multicolumn{3}{c}{PASCAL VOC}\\
    \cmidrule(lr){2-4}
    \cmidrule(lr){5-7}
    Loss                        & \ac{mIoU}~(\%)$\uparrow$    & \ac{Acc}~(\%)$\uparrow$ & \ac{mAP}~(\%)$\uparrow$ & \ac{mIoU}~(\%)$\uparrow$    & \ac{Acc}~(\%)$\uparrow$ & \ac{mAP}~(\%)$\uparrow$\\
    \midrule
    \multicolumn{1}{l}{Pretrained}         & 53.71 & 73.95 & 81.61 & 55.50 & \textbf{74.53} & 84.03\\
    \multicolumn{1}{l}{DINOv2}             & 54.90 & 74.76 & 82.15 & 53.07 & 72.14 & 83.23\\
    \multicolumn{1}{l}{w/ $\la$}           & 57.26 & 75.64 & 83.87 & 52.37 & 70.68 & 82.70\\
    \multicolumn{1}{l}{w/ $\lb$}           & 60.90 & 78.79 & 84.71 & 55.04 & 73.32 & 84.24\\
    \multicolumn{1}{l}{w/ $\la$ \& $\lb$}  & \textbf{61.63} & \textbf{78.86} & \textbf{85.40} & \textbf{55.70} & 73.30 & \textbf{84.73}\\
    \bottomrule
  \end{tabular}
\end{table}

For the unsupervised object discovery task, as shown in \cref{tab_discovery_largereg_continual}, 
we find that the models after continual learning, except for the model with $\lb$ alone, show improved performance compared to the pretrained model. 
We also find that the model with $\la$ alone has the highest performance in this task, while the original model and the model with $\la$ and $\lb$ perform at a similar level.

\begin{table}
  \caption{Unsupervised object discovery performance in continual learning setting. (Model: ViT-L/14-Reg)}
  \label{tab_discovery_largereg_continual}
  \centering
  \begin{tabular}{cccc}
    \toprule
    Loss                                   & VOC07 & VOC12 & COCO\\
    \midrule
    \multicolumn{1}{l}{Pretrained}         & 56.2  & 60.2  & 43.0\\
    \multicolumn{1}{l}{DINOv2}             & 57.0  & 61.4  & 44.6\\
    \multicolumn{1}{l}{w/ $\la$}           & \textbf{58.1}  & \textbf{62.9}  & \textbf{45.1}\\
    \multicolumn{1}{l}{w/ $\lb$}           & 53.5  & 59.1  & 42.5\\
    \multicolumn{1}{l}{w/ $\la$ \& $\lb$}  & 57.0  & 61.6  & 44.6\\
    \bottomrule
  \end{tabular}
\end{table}

\subsection{Loss weights analysis}
\label{subsec_abl_weights}
We investigate the effects of loss weights $(w_{\mathrm{in}}, w_{\mathrm{out}})$ in \cref{eq_totalloss} that balance the DINO loss $\mathcal{L}_{\mathrm{DINO}}$ and the two visual focusing loss functions $\la$ and $\lb$ on segmentation performance.
In this experiment, we train ViT-L/16 with $\la$, one with $\lb$, and one with $\la$ and $\lb$ with weights set to $0.05$ and $0.1$, respectively.
The results in \cref{tab_abl_weights} show that introducing each weight of $0.05$ leads to a substantial improvement in performance compared to the original DINOv2 model, 
whereas increasing the weights to $0.1$ slightly degrades performance compared to $0.05$. 
These findings indicate that simply increasing the weights of $\la$ and $\lb$ is not necessarily beneficial; 
rather, balancing between the visual focusing loss and the DINO loss is crucial for achieving optimal performance.
We conduct these analyses on training the ViT model without registers and find that the resulting loss weights also perform well with the ViT model incorporating registers. 
Therefore, we apply these weights in the experiments presented in the main results of our paper.

\begin{table}
  \caption{Zero-shot image segmentation performance across different loss weights.  (Model: ViT-L/16)}
  \label{tab_abl_weights}
  \centering
  \begin{tabular}{c@{\hspace{2pt}}c@{\hspace{2pt}}c@{\hspace{2pt}}c@{\hspace{2pt}}c@{\hspace{2pt}}c@{\hspace{2pt}}c}
    \toprule
    \multicolumn{1}{c}{Loss weights} & \multicolumn{3}{c}{ImageNet-Seg.} & \multicolumn{3}{c}{PASCAL VOC}\\
    \cmidrule(lr){2-4}
    \cmidrule(lr){5-7}
    $(w_{\mathrm{in}}, w_{\mathrm{out}})$                       & \ac{mIoU}~(\%)$\uparrow$    & \ac{Acc}~(\%)$\uparrow$ & \ac{mAP}~(\%)$\uparrow$ & \ac{mIoU}~(\%)$\uparrow$    & \ac{Acc}~(\%)$\uparrow$ & \ac{mAP}~(\%)$\uparrow$\\
    \midrule
    \multicolumn{1}{l}{$(0, 0)$}        & 51.94 & 72.04 & 81.23 & 49.66 & 68.81 & 82.19\\
    \addlinespace[2pt]
    \multicolumn{1}{l}{$(0.05, 0)$}     & 60.97 & 78.71 & 84.76 & 56.17 & 73.76 & 83.90\\
    \multicolumn{1}{l}{$(0.1, 0)$}      & 60.79 & 78.46 & 83.73 & 56.04 & 73.54 & 83.02\\
    \addlinespace[2pt]
    \multicolumn{1}{l}{$(0, 0.05)$}     & 64.10 & 81.26 & 85.87 & 60.39 & 77.35 & 85.91\\
    \multicolumn{1}{l}{$(0, 0.1)$}      & 62.94 & 80.55 & 85.73 & 58.87 & 76.39 & 85.35\\
    \addlinespace[2pt]
    \multicolumn{1}{l}{$(0.05, 0.05)$}  & 65.77 & 82.21 & 86.29 & 59.57 & 76.53 & 85.35\\
    \multicolumn{1}{l}{$(0.1, 0.1)$}    & 63.65 & 80.67 & 85.48 & 57.97 & 75.28 & 84.39\\
    \bottomrule
  \end{tabular}
\end{table}

\subsection{Analysis of the number of selected heads during training}
\label{subsec_numofheads}
The number of selected heads $h_{\mathrm{on}}$ during training is a hyperparameter of our training method. 
The number $h_{\mathrm{on}}$ refers to the number of heads selected during head selection in \ac{PHS}, and it affects training through changes in the output of the student model $P^{\mathrm{PHS}}_{\mathrm{s}}$. 
To investigate the effect of $h_{\mathrm{on}}$ on performance, we train ViT-L/16 by varying $h_{\mathrm{on}}$ across three settings: $\{2, 5, 8\}$. 
\Cref{tab_numheads_seg_large,tab_numheads_foir_large} show the effect of $h_{\mathrm{on}}$ on the performance for the zero-shot image segmentation task and the \ac{FOIR} task, respectively. 
For $h_{\mathrm{on}}=2$, the models achieve high performance on the segmentation task, whereas their performance on the \ac{FOIR} task is relatively low.
In the case of $h_{\mathrm{on}}=8$, while the model with $\la$ and $\lb$ exhibits high performance on the FOIR task, the performance is relatively low for other models and tasks.
Since $h_{\mathrm{on}}=5$ yields consistently high performance overall, we adopt $h_{\mathrm{on}}=5$.
We find that the resulting value $h_{\mathrm{on}}=5$ also performs well with the ViT model incorporating registers. 
Therefore, we apply $h_{\mathrm{on}}=5$ in the experiments presented in the main results of our paper.

\begin{table}
  \caption{Effect of $h_{\mathrm{on}}$ on zero-shot image segmentation performance. (Model: ViT-L/16)}
  \label{tab_numheads_seg_large}
  \centering
  \begin{tabular}{c@{\hspace{2pt}}c@{\hspace{2pt}}c@{\hspace{2pt}}c@{\hspace{2pt}}c@{\hspace{6pt}}c@{\hspace{2pt}}c@{\hspace{2pt}}c}
    \toprule
    \multicolumn{2}{c}{} & \multicolumn{3}{c}{ImageNet-Seg.} & \multicolumn{3}{c}{PASCAL VOC}\\
    \cmidrule(lr){3-5}
    \cmidrule(lr){6-8}
    $h_{\mathrm{on}}$ & Loss                       & \ac{mIoU}~(\%)$\uparrow$    & \ac{Acc}~(\%)$\uparrow$ & \ac{mAP}~(\%)$\uparrow$ & \ac{mIoU}~(\%)$\uparrow$    & \ac{Acc}~(\%)$\uparrow$ & \ac{mAP}~(\%)$\uparrow$\\
    \midrule
    --&\multicolumn{1}{l}{DINOv2}            & 51.94 & 72.04 & 81.23 & 49.66 & 68.81 & 82.19\\
    \addlinespace[2pt]
    2&\multicolumn{1}{l}{w/ $\la$}           & 64.43 & 81.12 & 85.83 & 58.53 & 75.51 & 84.91\\
    5&\multicolumn{1}{l}{w/ $\la$}           & 60.79 & 78.46 & 83.73 & 56.04 & 73.54 & 83.02\\
    8&\multicolumn{1}{l}{w/ $\la$}           & 58.06 & 76.50 & 82.59 & 53.94 & 72.01 & 82.23\\
    \addlinespace[2pt]
    2&\multicolumn{1}{l}{w/ $\lb$}           & 62.66 & 80.35 & 85.51 & 56.55 & 74.72 & 84.51\\
    5&\multicolumn{1}{l}{w/ $\lb$}           & 62.94 & 80.55 & 85.73 & 58.87 & 76.39 & 85.35\\
    8&\multicolumn{1}{l}{w/ $\lb$}           & 57.01 & 77.45 & 81.74 & 55.48 & 74.74 & 84.16\\
    \addlinespace[2pt]
    2&\multicolumn{1}{l}{w/ $\la$ \& $\lb$}  & 66.14 & 82.13 & 87.37 & 59.80 & 76.38 & 85.98\\
    5&\multicolumn{1}{l}{w/ $\la$ \& $\lb$}  & 65.77 & 82.21 & 86.29 & 59.57 & 76.53 & 85.35\\
    8&\multicolumn{1}{l}{w/ $\la$ \& $\lb$}  & 63.87 & 81.01 & 85.90 & 59.08 & 76.28 & 85.24\\
    \bottomrule
  \end{tabular}
\end{table}

\begin{table}
    \caption{Focus-Oriented Image Retrieval performance. (Model: ViT-L/16, Retrieval scheme: Visual prompt per object)}
    \label{tab_numheads_foir_large}
    \centering
    \begin{tabular}{c@{\hspace{2pt}}c@{\hspace{2pt}}c@{\hspace{2pt}}c}
        \toprule
         &  & COCO & PASCAL VOC\\
        \cmidrule(lr){3-3}
        \cmidrule(lr){4-4}
        $h_{\mathrm{on}}$ & Loss & $\mathrm{MP@}100~(\mathrm{\%})\uparrow$ & $\mathrm{MP@}100~(\mathrm{\%})\uparrow$\\
        \midrule
        --&\multicolumn{1}{l}{DINOv2}            & 55.1 & 73.9\\
        \addlinespace[2pt]
        2&\multicolumn{1}{l}{w/ $\la$}           & 54.9 & 75.2\\
        5&\multicolumn{1}{l}{w/ $\la$}           & 57.5 & 78.1\\
        8&\multicolumn{1}{l}{w/ $\la$}           & 56.7 & 77.6\\
        \addlinespace[2pt]
        2&\multicolumn{1}{l}{w/ $\lb$}           & 49.4 & 69.5\\
        5&\multicolumn{1}{l}{w/ $\lb$}           & 55.8 & 76.0\\
        8&\multicolumn{1}{l}{w/ $\lb$}           & 52.8 & 72.3\\
        \addlinespace[2pt]
        2&\multicolumn{1}{l}{w/ $\la$ \& $\lb$}  & 54.2 & 74.0\\
        5&\multicolumn{1}{l}{w/ $\la$ \& $\lb$}  & 55.6 & 75.4\\
        8&\multicolumn{1}{l}{w/ $\la$ \& $\lb$}  & 56.8 & 77.9\\
        \bottomrule
    \end{tabular}
\end{table}

\subsection{Additional qualitative results of attention maps}
\label{subsec_additional_attention}
\begin{figure}
    \centering
    \includegraphics[width=\linewidth]{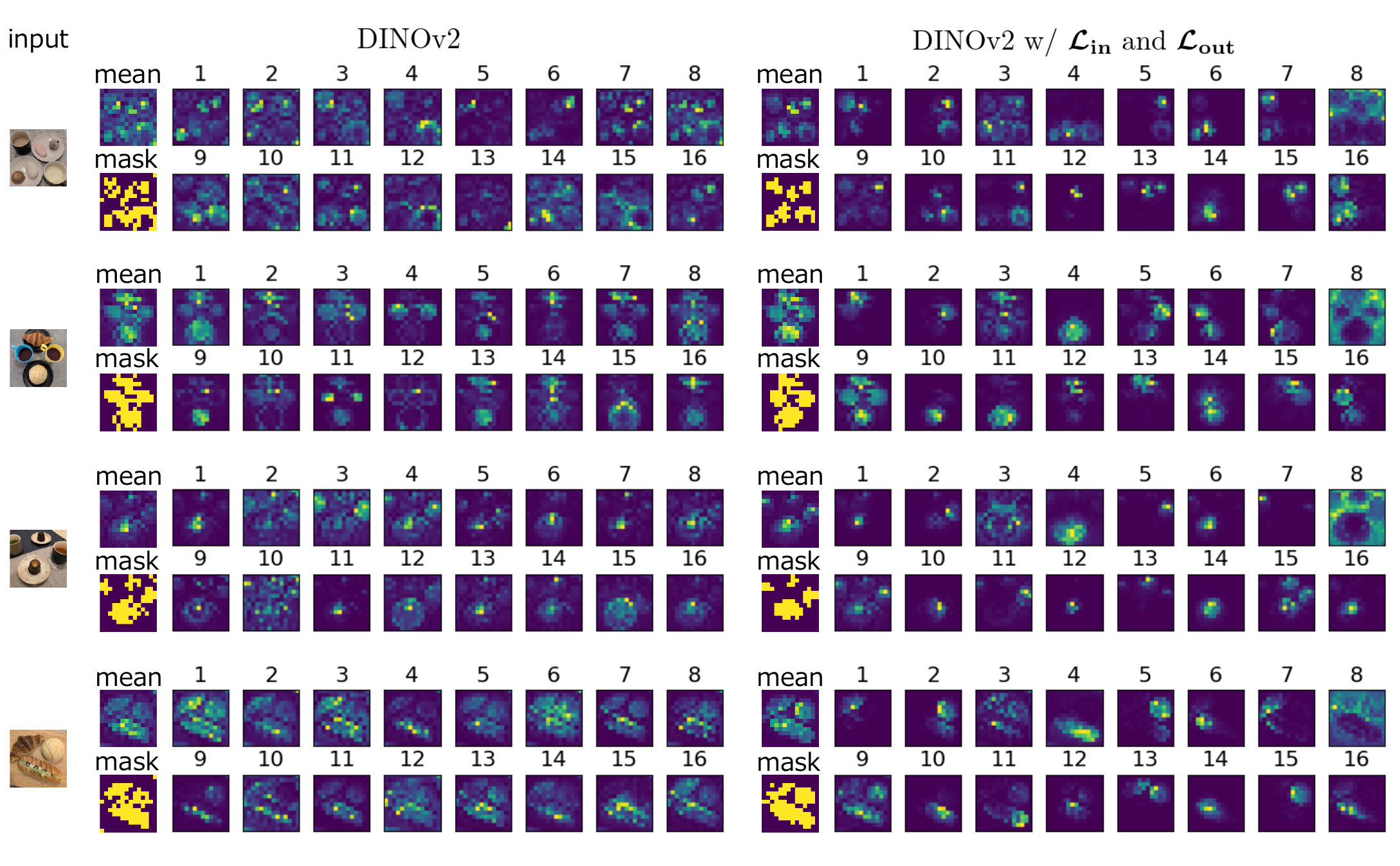}
    \caption{Qualitative comparison of attention maps for the original DINOv2 model and the model with $\la$ and $\lb$ on example inputs. (Model: ViT-L/16-Reg)}
    \label{fig_attention_additional_example}
\end{figure}

\Cref{fig_attention_additional_example} shows attention maps on additional example input images, similar to those in \cref{fig_attention_example}.
It is confirmed that the properties of attention maps discussed in \cref{sebsec_attention} are consistent across different examples.

\subsection{Entropy analysis of attention}
\label{subsec_entropy}
To complement our qualitative visualizations of attention maps in \cref{fig_attention_average_imagenetseg_largereg}, 
we additionally conduct a quantitative analysis of spatial attention head specialization. 
Specifically, we compute the Shannon entropy of the attention distribution underlying the attention maps. 
\Cref{tab_entropy} shows the following two metrics: (1) Mean Per-Head Entropy: the entropy is first computed separately for each head 
and then averaged across heads, and (2) Entropy of Head-Averaged Attention: the attention distributions are first averaged across heads, after which the entropy is computed. 
These entropies are normalized, such that a uniform attention distribution has the maximum value of $1$. 
We use models evaluated in \cref{fig_attention_average_imagenetseg_largereg} and report averages over the validation set of ImageNet-1k. 
Our $\la$ \& $\lb$ model exhibits lower mean per-head entropy than the original DINOv2 model but high entropy after averaging attention across heads, 
indicating that individual heads produce spatially concentrated attention, while their aggregation covers a broad range of image regions. 
This result provides quantitative evidence of spatial specialization and diversity across attention heads. 

\begin{table}
  \caption{Entropy-based evaluation of spatial attention head specialization and diversity}
  \label{tab_entropy}
  \centering
  \begin{tabular}{c@{\hspace{2pt}}c@{\hspace{6pt}}c}
    \toprule
    Loss                        & Mean Per-Head Entropy   & Entropy of Head-Averaged Attention\\
    \midrule
    \multicolumn{1}{l}{DINOv2}             & 0.8841 & 0.9265\\
    \multicolumn{1}{l}{w/ $\la$ \& $\lb$}  & 0.7492 & 0.9166\\
    \bottomrule
  \end{tabular}
\end{table}

\subsection{Supervised semantic segmentation}
\label{subsec_supervised_seg}
We evaluate supervised semantic segmentation using linear probing, following the setting of \cite{dinov3}.
We conduct the evaluations on ADE20k~\citep{Zhou_2017_CVPR}, Cityscapes~\citep{cityscapes}, and PASCAL VOC 2012 (VOC)~\citep{everingham2015pascalvoc}.
For VOC, we train the models on the trainaug dataset~\citep{SBD}. 
We perform a hyperparameter sweep varying the learning rate over $\{1\times 10^{-4},~3\times 10^{-4},~1\times 10^{-3}\}$ and 
weight decay over $\{1\times 10^{-4},~1\times 10^{-3}\}$. 

\Cref{tab_supervised_seg} compares the supervised semantic segmentation performance of our $\la$ \& $\lb$ model with that of the original DINOv2 model. 
In terms of model scaling, for both DINOv2 and our model, the Large variant outperforms the Base variant in supervised semantic segmentation. 
However, in the zero-shot segmentation setting of \cref{tab_seg_largereg,tab_seg_baseandgiantreg}, 
we observe the opposite trend, with the Base variant outperforming the Large variant. 
This trend is consistent with both previous work~\citep{stegmuller2025a} and our additional evaluation of the publicly available pretrained DINOv2 models 
under the same zero-shot segmentation protocol (\cref{tab_seg_openmodel_size}). 

Regarding method comparison in \cref{tab_supervised_seg}, our $\la$ \& $\lb$ model consistently outperforms the original DINOv2 model 
across all three datasets for the Base model size. 
For the Large model size, the $\la$ \& $\lb$ model demonstrates superior performance on Cityscapes, while the original DINOv2 model achieves better results 
on ADE20k and VOC. 
Nevertheless, the performance differences between the two models are small in the supervised semantic segmentation setting, with both achieving competitive results 
and exhibiting no clear performance advantage over one another. 
These findings clarify that the distinctive property of our model, namely spatial attention head specialization, 
enables the direct extraction of high-quality features from attention maps in a zero-shot setting 
while maintaining competitive performance on the supervised downstream task of semantic segmentation. 

\begin{table}
  \caption{Supervised semantic segmentation performance using linear probing. 
  The evaluation metric is mIoU (\%). (Model: ViT-\{Size\}/16-Reg)}
  \label{tab_supervised_seg}
  \centering
  \begin{tabular}{c@{\hspace{2pt}}c@{\hspace{2pt}}c@{\hspace{4pt}}c@{\hspace{4pt}}c}
    \toprule
    Size & Loss                        & ADE20k    & Cityscapes & VOC\\
    \midrule
    \multirow{2}{*}{B}  &\multicolumn{1}{l}{DINOv2}            & 30.80 & 54.63 & 66.82\\
                        &\multicolumn{1}{l}{w/ $\la$ \& $\lb$} & \textbf{31.14} & \textbf{55.30} & \textbf{68.66}\\
    \midrule
    \multirow{2}{*}{L}  &\multicolumn{1}{l}{DINOv2}            & \textbf{43.15} & 64.25 & \textbf{81.13}\\
                        &\multicolumn{1}{l}{w/ $\la$ \& $\lb$} & 42.17 & \textbf{64.82} & 80.89\\
    \bottomrule
  \end{tabular}
\end{table}

\begin{table}
  \caption{Zero-shot segmentation performance of publicly available DINOv2 pretrained models~\citep{darcet2023vision}. (Model: ViT-\{Size\}/16-Reg, Dataset: ImageNet-Seg.)}
  \label{tab_seg_openmodel_size}
  \centering
  \begin{tabular}{c@{\hspace{6pt}}c@{\hspace{2pt}}c@{\hspace{4pt}}c}
    \toprule
    Size                       & \ac{mIoU}~(\%)$\uparrow$    & \ac{Acc}~(\%)$\uparrow$ & \ac{mAP}~(\%)$\uparrow$\\
    \midrule
    B  & 65.63 & 82.50 & 85.69\\
    \midrule
    L  & 53.71 & 73.95 & 81.61\\
    \midrule
    g  & 55.56 & 75.47 & 82.41\\
    \bottomrule
  \end{tabular}
\end{table}

\begin{table}[!t]
  \caption{Comparison of the original model, our model in the main paper, and an additional model with PHS applied to the third-to-last attention block. 
  Zero-shot segmentation is evaluated on ImageNet-Seg. (Model: ViT-L/16-Reg)}
  \label{tab_phs_block}
  \centering
  \begin{tabular}{c@{\hspace{2pt}}c@{\hspace{2pt}}c@{\hspace{2pt}}c@{\hspace{2pt}}c@{\hspace{2pt}}c}
    \toprule
    \multicolumn{1}{c}{} & \multicolumn{3}{c}{Zero-shot segmentation} & \multicolumn{1}{c}{} & \multicolumn{1}{c}{}\\
    \cmidrule(lr){2-4}
    Loss                        & \ac{mIoU}~(\%)$\uparrow$    & \ac{Acc}~(\%)$\uparrow$ & \ac{mAP}~(\%)$\uparrow$ & k-NN~(\%) & Linear~(\%)\\
    \midrule
    \multicolumn{1}{l}{DINOv2}                                     & 63.67 & 80.69 & 85.39                   & \textbf{81.7} & \textbf{82.6}\\
    \multicolumn{1}{l}{w/ $\la$ \& $\lb$,} last block PHS          & 67.29 & 83.20 & \textbf{86.43}          & 80.3 & 81.8\\
    \multicolumn{1}{l}{w/ $\la$ \& $\lb$,} third-to-last block PHS    & \textbf{67.85} & \textbf{84.09} & 86.35 & 81.0 & 82.2\\
    \bottomrule
  \end{tabular}
\end{table}

\subsection{Applying head selection beyond the last attention block}
\label{subsec_phs_block}
While our main experiments apply the head selection method to the last attention block of the student model, 
we further explore its application to other blocks and discuss the trade-off between the spatial task performance and classification performance.
Since attention blocks can be interpreted as integrating information from the \verb|[CLS]| token and patch tokens, 
we hypothesize that retaining standard attention blocks after the PHS-modified block allows spatially specialized attention information 
to propagate to the \verb|[CLS]| representation used for classification. 

As shown in \cref{tab_phs_block}, applying PHS to the third-to-last attention block with $h_{\mathrm{on}} = 2$ preserves high zero-shot segmentation 
performance while reducing the accuracy gaps relative to the original DINOv2 model to $0.7\%$ for k-NN classification and $0.4\%$ for linear probing. 
Although fully closing the classification gap remains an important direction for future work, we believe these additional results highlight the promise of our approach.


\end{document}